\documentclass{article}
\usepackage{ijcai26}

\usepackage{times}
\usepackage{soul}
\usepackage{url}
\usepackage[utf8]{inputenc}
\usepackage[small]{caption}
\usepackage{graphicx}
\usepackage{amsmath}
\usepackage{amsthm}
\usepackage{booktabs}
\usepackage{algorithm}
\usepackage{algorithmic}
\usepackage[switch]{lineno}

\usepackage{multirow}
\usepackage{amsfonts}

\title{Learning to Route LLMs from Implicit Cost-Performance Preferences via Meta-Learning}

\author{
Jiahao Zeng$^1$
\and
Ming Tang$^2$
\And
Ningning Ding$^{1}$\thanks{Corresponding author.}\\
\affiliations
$^1$Hong Kong University of Science and Technology (Guangzhou)\\
$^2$Southern University of Science and Technology\\
\emails
jzeng110@connect.hkust-gz.edu.cn,
tangm3@sustech.edu.cn,
ningningding@hkust-gz.edu.cn
}

\begin{document}

\maketitle

\begin{abstract}
    Large language models (LLMs) present a trade-off between performance and cost, where more powerful models incur greater expense. LLM routing aims to mitigate expenses while maintaining performance by sending queries to the most suitable model. However, existing methods cannot perform well for different user cost-performance preferences. To address this gap, we introduce a novel perceptive LLM routing paradigm for personalized and user-centric cost-performance optimization, which efficiently learns users' implicit preferences through little interaction. To handle the challenge of heterogeneous user needs, we formulate preference profiles as a set of distinct tasks in contextual bandit and propose MetaRouter, a meta-learning framework designed for preference-aware LLM routing. Experimental results show that MetaRouter outperforms strong baselines on both in-distribution and out-of-distribution tasks. Furthermore, it exhibits high efficiency in learning user preferences, robustness to changes in the routable LLMs, and scalability to multi-model routing.
\end{abstract}

\section{Introduction}

The rapid evolution of Large Language Models (LLMs) like GPT series \cite{OpenAIResearch} marks a significant breakthrough in artificial intelligence. These models excel at a wide array of natural language processing tasks, including complex reasoning and sophisticated problem-solving \cite{sparks,cot}. However, their immense power comes at a cost. The substantial computational and memory requirements of these state-of-the-art LLMs necessitate cloud-based deployment and expensive API access. In response to these challenges, a new class of smaller and more efficient LLMs has emerged \cite{llama2023,phi3}. These models are specifically designed to operate on edge devices such as personal computers. This new paradigm offers compelling advantages, including low-cost deployment, reduced latency, and enhanced data privacy. However, it also introduces a critical trade-off: while constantly improving, the performance of these smaller models still lags behind the large LLMs \cite{hybridllm}.

To balance performance and cost, a common and effective strategy is to employ a hybrid system that balances the use of large and powerful models with smaller and more cost-effective ones. Some approaches employ a cascade method which sequentially queries different LLMs until a reliable response is found \cite{frugalgpt,automix}. However, this usually needs to query multiple times and leads to high latency. A more promising solution is to use routing methods, which direct each query to a single and optimal LLM. Similar to a small model, the router can be deployed on edge devices. It can route simple queries to a local small LLM and only use an API call for difficult queries. This strategy provides a flexible way to maximize LLM power for various budgets and quality needs.


\begin{figure}[t]
\centering
\includegraphics[width=0.9\columnwidth]{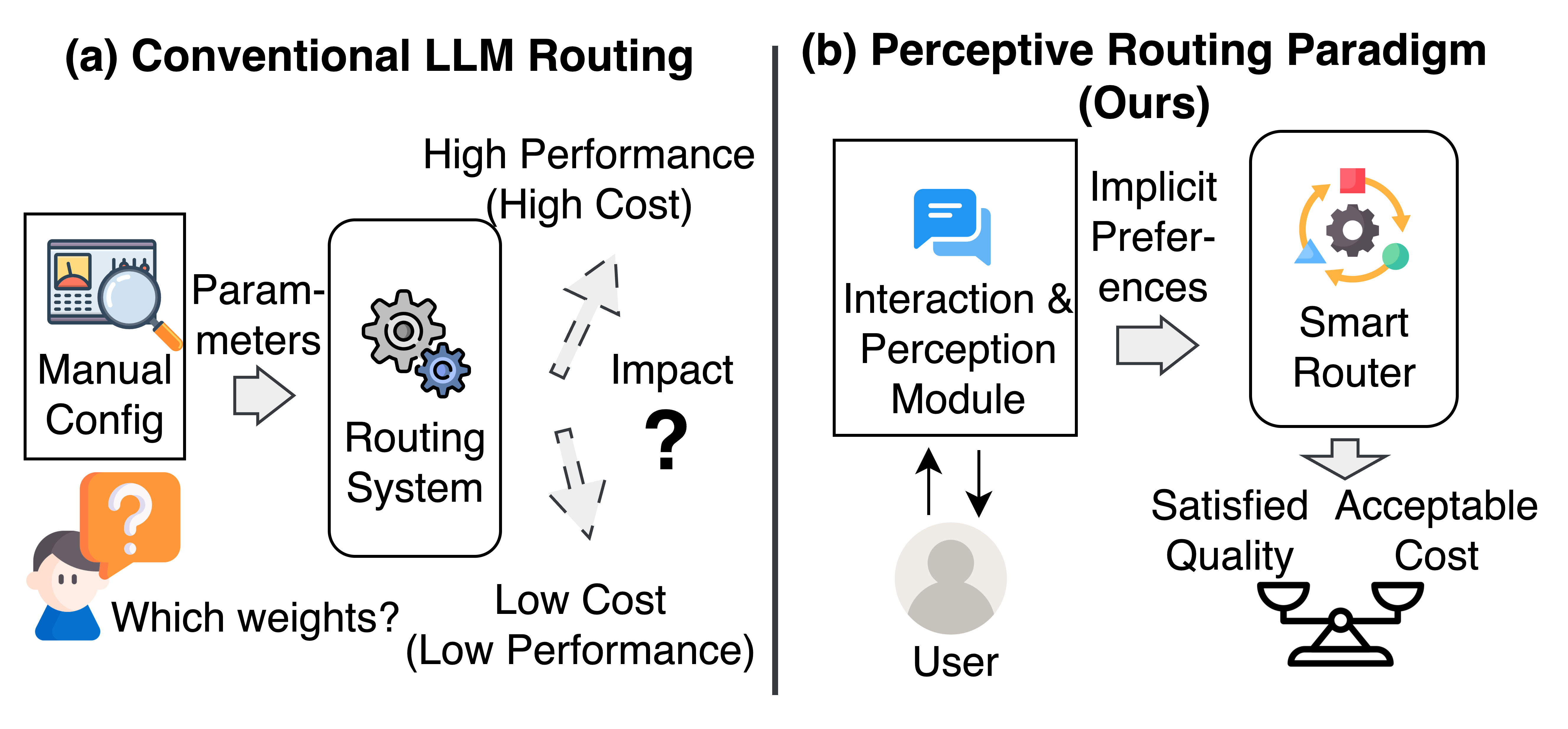} 
\caption{Perceptive LLM routing paradigm.}
\label{fig:routing comparison}
\end{figure}

Existing research has explored various methods for routing queries. \cite{hybridllm} used a router that assigns queries to a small or large model based on the predicted query difficulty and the desired quality level. \cite{routellm} proposed several efficient router models and developed a training framework that leverages human preference data. \cite{graphrouter} introduced a graph-based framework that utilized contextual information among tasks, queries, and LLMs to improve selection. However, a significant limitation of existing approaches is their reliance on manually configured parameters to strike a balance between performance and cost, as shown in Figure~\ref{fig:routing comparison}(a). Moreover, some advanced methods like \cite{graphrouter} that attempted to learn these trade-offs require retraining a new model for each distinct user preference, which is both inefficient and computationally expensive.

To address these limitations, we introduce a perceptive LLM routing paradigm. As shown in Figure~\ref{fig:routing comparison}(b), the router learns users' implicit preferences through interaction. However, realizing this paradigm presents two core challenges: (a) \emph{Feedback Collection}: How to collect user feedback unobtrusively while ensuring it fully reflects users' preferences? (b) \emph{Preference Integration}: How to infer preferences from this feedback and incorporate them into routing decisions?

In this paper, we propose MetaRouter, an end-to-end meta-learning framework for preference-aware LLM routing. We formulate the routing decision as a contextual bandit problem and treat distinct preferences as different tasks. During the meta-training phase, we train the router across diverse preference profiles to enable rapid adaptation. Specifically, our method initiates an adaptation phase where users provide pairwise comparisons of LLM responses. From this comparative feedback, we infer a latent preference representation that captures their cost-performance trade-offs. This representation is combined with users' query as input for the routing policy, allowing the system to intelligently select the optimal model for each query. By conditioning decisions on both the queries and learned preferences, our approach offers a highly personalized experience. Overall, our contributions include:
\begin{itemize}
    \item \emph{Novel Paradigm for LLM Routing.} We introduce \emph{perceptive} LLM routing, a novel paradigm that goes beyond traditional \emph{manual} configurations or model retraining. Our approach learns users' implicit preferences for cost–performance trade-offs through interaction, enabling the system to adaptively select the most suitable LLM for each query based on individual user needs.
    \item \emph{Meta-Learning Framework for Preference-Awareness.} We introduce MetaRouter, a meta-learning-based approach that solves the preference-aware routing problem. To our knowledge, this is the first work to build LLM router based on meta-learning, providing insights into advancing more intelligent LLM systems.
    \item \emph{Comprehensive Evaluation. } We validate MetaRouter on both in-distribution and out-of-distribution datasets. The results show that our method outperforms strong baselines across multiple performance metrics. Experiments also show its efficiency in learning user preferences, robustness to changes in the routable LLMs, and scalability to multi-model routing.
\end{itemize}

\section{Related Work}
\paragraph{Hybrid LLM.}
To mitigate the expenses of using LLMs, an effective strategy is to implement a hybrid system that balances the use of large and small models. Early cascade methods \cite{frugalgpt,automix} query models sequentially from cheapest to most expensive, but this can introduce significant latency. In contrast, routing methods \cite{hybridllm,routellm,graphrouter} provide a more promising solution by directly routing queries to the most suitable model. For instance, \cite{hybridllm} used a router to assign queries based on predicted difficulty and desired quality. \cite{routellm} proposed several efficient router models and developed a training framework leveraging human preference data to enhance performance. \cite{graphrouter} introduced GraphRouter, a framework that utilized contextual information between tasks, queries, and LLMs to improve LLM selection. Despite these advances, a significant limitation of existing approaches is their reliance on manually configured parameters to balance performance and cost, which is not user-friendly. Moreover, solutions like GraphRouter \cite{graphrouter} require retraining a new model for each distinct user preference, which is both inefficient and computationally expensive. In this work, we introduce a perceptive LLM routing paradigm to learn users' preferences.

\paragraph{Meta Learning for Bandits.} 
Meta-learning represents a paradigm that enables the model to learn a learning strategy itself. Instead of training a model to master a single task, meta-learning algorithms are trained on a distribution of tasks. Applying this principle to reinforcement learning gives rise to Meta-Reinforcement Learning (Meta-RL), which addresses poor generalization in traditional RL. Meta-RL aims to learn policies that can adapt to new tasks with minimal data. While extensively studied in general RL settings \cite{surveymetarl}, meta-learning in the bandit framework remains less explored. Prior work has investigated meta-learning for adversarial bandit feedback \cite{AdversarialBanditAlgorithms}, learning policies in Bayesian bandits \cite{DifferentiableMeta-Learning}, and developing exploration strategies \cite{Meta-LearningEffectiveExplorationStrategies}. However, these methods are not applicable to our problem due to different settings. In this work, we frame the preference-aware LLM routing problem from a meta-learning perspective and propose an algorithm that learns a routing policy capable of quickly adapting to users' preferences.

\section{Problem Formulation}

\subsection{LLM Routing}

We use $X$ and $O$ to denote the user query space and the LLM answer space, respectively. Let $\mathcal{M} = \{M_1, M_2, \dots, M_K\}$ represent a set of $K$ candidate LLMs available for routing. While our proposed framework supports routing among multiple models (as validated in Section \ref{sec:gen_scale}), we focus our problem formulation on the most common binary scenario ($K=2$) for clarity and comparativity \cite{hybridllm,routellm}. Specifically, we consider two distinct models: a large LLM, denoted as $L:X\rightarrow O$, and a small LLM, denoted as $S:X\rightarrow O$. Large models refer to state-of-the-art models such as the latest GPT series \cite{OpenAIResearch}, which are often accessed via APIs. These models offer superior performance but typically incur higher latency and monetary costs. Conversely, small models refer to lightweight models with fewer parameters that can be deployed locally, such as the Phi family \cite{phi3}. They have lower costs and weaker performance, but can achieve performance close to that of large models on many simple tasks. 

LLM routing can be modeled as a sequential decision-making problem, which can be addressed using a contextual bandit framework. Specifically, at timestep $t$, a routing policy $\pi$ observes a state (query $x_t$) and selects an action $a_t \in \{S, L\}$, corresponding to the choice of model. To avoid ambiguity with ``context" as used in task inference methods (introduced in the next subsection), we exclusively use the term ``state" to refer to the contextual information in the bandit formulation. The reward for routing query $x$ to model $a$ is denoted by $R(x, a)$, a function that depends on user utility. The performance of the routing policy can be measured by:

\begin{equation}
\label{eq:original_obj}
J'(\pi) = \mathbb{E}_{x\sim X}\left[ R(x,a) \mid a \sim \pi(x) \right].
\end{equation}

Model selection often involves a trade-off between performance and cost, and users have different preferences for the trade-off. To capture this, we define a general score function $s(o)$ to represent the user utility of an LLM-generated response $o$. Crucially, our framework is agnostic to the specific form of $s(o)$. This function constructs the reward that guides the policy's learning, as detailed in the next section.

User utility depends on both performance and cost of response $o$. Specifically, we use $q(o)$ to represent the performance of a response $o$, obtainable through evaluation methods such as BART score \cite{hybridllm} or a LLM judge \cite{routellm}. Concurrently, we use $p(o)$ to denote the inference cost of $o$, which is based on the token count and price. During the meta-training phase (detailed in Section~\ref{sec:meta-training}), we can generate a diverse set of synthetic tasks (i.e., simulated user preferences) by varying the preference parameters (e.g., trade-off weights) inherent to the definition of $s(o)$. For example, the most common form is $s(o)=w\cdot q(o)-(1-w)\cdot p(o)$ \cite{graphrouter,avengerspro}. Note that to reduce the influence of their scales, both $p(o)$ and $q(o)$ are normalized.

\subsection{Few-shot Preference Learning}
Current methods \cite{hybridllm,routellm,avengerspro} require users to manually configure parameters to adjust their preferences for cost and performance. However, this paradigm is unintuitive and presents usability challenges. Primarily, users often do not know the actual impact of an abstract parameter. Furthermore, this paradigm assumes that users have a clear understanding of the optimal parameter for their specific needs. Consequently, this approach can lead to suboptimal outcomes for users. In this work, we propose a method to infer user preferences from their feedback.

Motivated by the success of task inference methods in Meta-RL \cite{surveymetarl,PEARL}, we formulate different preferences as distinct tasks and aim to identify the ``task" by collecting users' preference contexts. However, unlike standard Meta-RL, which leverages continuous environmental interactions (states, actions, and rewards) to infer tasks, the router cannot rely on constant user interaction, because collecting persistent preference feedback would severely degrade the user experience. Consequently, we decouple the inputs to the task inference method from the outputs of the policy. The task inference method is conditioned only on preference contexts collected during a preliminary adaptation process. In this process, we return responses from both models for a query and require the user to choose the better one by considering prices and performance. This feedback constitutes a preference context $c$. For example, if the user feels that the small model $S$ is better for query $x$, then $c = (x, S \succ L)$. The collected contexts are denoted by $c_{1:N}$.

Regarding the task inference, we represent each task $\mathcal{T}$, corresponding to a unique preference profile, with a $d$-dimensional latent task variable $z \in Z$. The policy $\pi_{\theta}(a|x,z)$, parameterized by $\theta$, is conditioned on the task variable $z$, enabling end-to-end learning of the representation $z$ alongside the policy, which aims to differentiate between task specifications. Additionally, we introduce an inference network (context encoder) $\mathcal{E}_\phi(z|c_{1:N})$, parameterized by $\phi$, to infer the task variable $z$ from preference contexts $c_{1:N}$. The policy's objective is to maximize the expected reward:

\begin{equation}
\label{eq:z_obj}
J(\pi_{\theta})= \mathbb{E}_{x\sim X, z \sim \mathcal{E}_{\phi}} \big[ R(x,a) \mid a \sim \pi_\theta(x,z) \big].
\end{equation}

In this case, our problem presents several challenges. First, it is hard to capture implicitly diverse user preferences through explicit tasks. Second, acquiring preference contexts from users' limited feedback results in a challenging data-scarce learning environment. Finally, it is difficult to learn an effective policy for a contextual bandit problem, which is only one-step RL. We will address them in the next section.

\section{Methodology}
\begin{figure*}[t]
    \centering
    \includegraphics[width=0.9\textwidth]{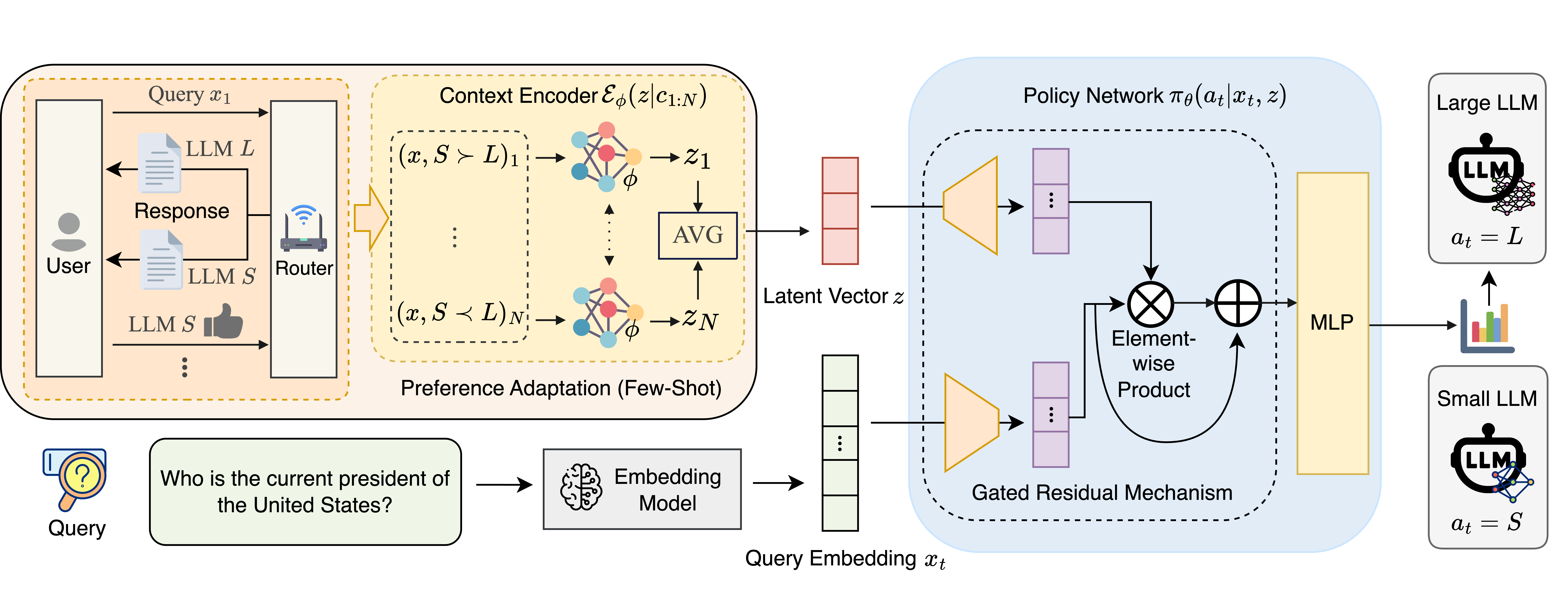}

    \caption{Overview of MetaRouter. MetaRouter consists of two components: a context encoder $\mathcal{E}_\phi$ that learns user preferences from their interactions, and a policy network $\pi_\theta$ that routes their queries based on those perceived preferences.}
    \label{fig:overview_twostage}
    \end{figure*}

As shown in Figure \ref{fig:overview_twostage}, MetaRouter comprises two core components: a context encoder $\mathcal{E}_{\phi}$ and a policy network $\pi_\theta$. The context encoder $\mathcal{E}_{\phi}$ collects users' preference contexts during an initial adaptation phase to infer the latent task variable $z$. Subsequently, each query will be input into the policy network $\pi_\theta$ together with $z$ for intelligent routing. The following subsections will detail each component and describe how they are trained jointly through meta-training.

\subsection{Context Encoder}
To adapt to diverse user preferences, our router employs a context encoder $\mathcal{E}_{\phi}(z|c_{1:N})$. This network infers a latent context vector $z$ which encapsulates users' preference information. This inference is conditioned on a collection of feedback contexts $c_{1:N}$ gathered during an adaptation phase.

A key challenge in personalization is designing a feedback mechanism that is both informative for the router and unobtrusive for the user. To achieve this, we employ a lightweight yet effective feedback system based on pairwise comparisons. During the adaptation process, the system presents the user with responses from both models. The user simply chooses the preferred one, considering both performance and price. This feedback constitutes a context $c$. This approach is cognitively less demanding than asking for absolute scores and provides a direct signal of the user's latent trade-offs.

While the feedback contexts $c_{1:N}$ are collected sequentially, we assume the underlying user preference profile remains stationary during adaptation. Therefore, the final representation $z$ must be independent of the arbitrary order in which feedback is received. To satisfy this property, we formulate $\mathcal{E}_{\phi}(z|c_{1:N})$ using a permutation-invariant function:
\begin{equation}
\label{eq:q}
\mathcal{E}_{\phi}(z|c_{1:N}) = \frac{1}{N} \sum_{n=1}^{N} f_{\phi}(c_n),
\end{equation}
where the function $f$ is a neural network which outputs a $d$-dimensional vector. The parameters of $\mathcal{E}_{\phi}$ are learned during the meta-training phase and remain fixed during inference. This deterministic task variable $z$ offers stable and clear guidance to the policy network.

\subsection{Routing Policy}
\subsubsection{Optimization Objective}
For the routing policy $\pi$, we employ the policy gradient (PG) method for optimization. Recall that we model the LLM routing problem as a contextual bandit, which is a one-step RL problem. At each step, the policy observes a state (the query $x$) and makes a decision (selects a model $a$) to maximize an immediate reward. Given this ``one-step" nature, PG is sufficient because it does not need to account for long-term consequences or a sequence of states. 

However, standard PG methods can suffer from premature convergence, where the policy becomes deterministic too quickly, failing to sufficiently explore the capabilities of different LLMs. To mitigate this and encourage exploration, we incorporate an entropy regularization term into the objective function. The overall objective is maximizing the expected reward augmented by the entropy of the policy:
\begin{equation}
    J(\theta) = \mathbb{E}_{x, z, a} [R(x, a)] + \beta \mathbb{E}_{x, z} [H(\pi(\cdot|x, z))],
\end{equation}
where $H(\pi(\cdot|x, z))$ is the entropy of the policy distribution, and $\beta$ is a hyperparameter controlling regularization. This ensures the router maintains stochasticity during early training, preventing collapse to a trivial solution (e.g., always selecting the largest model).

\subsubsection{Reward Design}
To ensure scalability to multi-model scenarios and stabilize the training process, we adopt a baseline subtraction technique. Specifically, we define the reward as the deviation of the selected model's utility $s(o_a)$ from the average utility of all candidate models for the given query $x$. Furthermore, to normalize the reward scale and mitigate the impact of outliers, we apply a hyperbolic tangent function. The formulated reward function is:
\begin{equation}
\label{eq:reward}
R(x,a) = \tanh\left(\lambda \left( s(o_a) - \bar{s}(x) \right) \right),
\end{equation}
where $\bar{s}(x) = \frac{1}{|\mathcal{A}|}\sum_{a' \in \mathcal{A}} s(o_{a'})$ represents the average utility of the candidate models, and $\lambda$ is a scaling factor controlling the sensitivity of the reward. This formulation encourages the policy to select models that perform above the average, providing a robust signal for training.

\subsubsection{Network Architecture}
Conventional task inference approaches typically rely on concatenating the observation with the latent task variable. However, this strategy is suboptimal in our routing problem due to the large dimensionality disparity between the dense query embedding (e.g., 768 dimensions) and the compact task variable $z$ (e.g., 10 dimensions). Such direct concatenation risks overshadowing the information in $z$. Furthermore, simple concatenation provides only implicit interaction, limiting the ability of $z$ to explicitly modulate the query representation. To overcome these limitations, we introduce a Gated Residual Mechanism, as illustrated in the right component of Figure \ref{fig:overview_twostage} and detailed in Supplementary Material.

Our architecture processes the inputs through two parallel branches. Specifically, we utilize an observation encoder (parameterized as an MLP) to project the query embedding into a feature representation $h_{obs}$. In parallel, the latent variable $z$ is processed by a gating network to produce a modulation vector $g$. This process is formally defined as:
\begin{equation}
    g = \tanh(\text{MLP}_{gate}(z)),
\end{equation}
where $\text{MLP}_{gate}$ denotes a multi-layer perceptron projecting $z$ to the same dimension as $h_{obs}$, and the $\tanh$ activation constrains the gate values to the range $[-1, 1]$, allowing for both feature suppression and enhancement. Crucially, instead of simple multiplication, we employ a residual connection to stabilize the training:
\begin{equation}
    h_{mod} = h_{obs} + (h_{obs} \odot g).
\end{equation}
where $\odot$ denotes the Hadamard product. This formulation allows the user preference $z$ to selectively enhance (positive $g$) or suppress (negative $g$) specific dimensions of the query while preserving the original query information flow. Finally, $h_{mod}$ is fed into the MLP to output the routing probabilities.

\subsection{Meta-training}
\label{sec:meta-training}
The central goal of meta-training is to equip MetaRouter with the ability to rapidly adapt to diverse user preferences from minimal feedback. This is achieved by training the context encoder $\mathcal{E}_\phi$ and the routing policy $\pi_\theta$ across a wide distribution of simulated tasks, where each task represents a unique user preference profile, as detailed in Algorithm \ref{alg:meta-training}.

\begin{algorithm}[t]
    \caption{Meta-training}
    \label{alg:meta-training}
      \textbf{Input:} Batch of tasks $\mathbb{T}=\{\mathcal{T}_{i}\}_{i=1...T}$\\
      \textbf{Parameter:} Learning rates $\alpha_{1}, \alpha_{2}$, Noise scale $\sigma$, Noise probability $p_{noise}$\\
      \textbf{Output:} $\pi_{\theta}, \mathcal{E}_{\phi}$
    \begin{algorithmic}[1]
    \FOR{iteration in training iterations}
        \STATE Sample a batch of training tasks $\mathbb{T}_{train}$.
        \FOR{each $\mathcal{T}^{i} \in \mathbb{T}_{train}$}
            \STATE Sample contexts $c_{1:N}^{i} \sim S_{c}(X)$ and batch of queries $B^{i} \sim X$.
            \STATE $\tilde{B}^{i} \leftarrow \{ x + b_x \cdot \epsilon \mid x \in B^{i} \}$, where $b_x \sim \text{Bernoulli}(p_{noise})$ and $\epsilon \sim \mathcal{N}(0, \sigma^2 I)$.
            \STATE Sample $z \sim \mathcal{E}_{\phi}(z|c_{1:N}^{i})$.
            \STATE Calculate objective: $J_{\theta}^{i} = J_{\theta}(\tilde{B}^{i}, z)$.
        \ENDFOR
        \STATE $\phi \leftarrow \phi + \alpha_{1} \nabla_{\phi} \sum_{i} J_{\theta}^{i}$
        \STATE $\theta \leftarrow \theta + \alpha_{2} \nabla_{\theta} \sum_{i} J_{\theta}^{i}$
    \ENDFOR
    \end{algorithmic}
    \end{algorithm}

In each training iteration, we uniformly sample a batch of tasks $\mathbb{T}_{train}$ (line 2). For each sampled task $\mathcal{T}^{i}$, we generate a set of preference contexts $c_{1:N}^{i}$ and a batch of queries $B^{i}$ (line 4). The contexts are derived by sampler $S_c$ which applies the specific task's scoring function to LLM outputs, simulating the pairwise feedback we would get from a real user with that preference profile.

To enhance the router's robustness against input variations, we employ a noise injection strategy (line 5). Formally, for each query $x$ in the batch $B^{i}$, we independently sample a binary mask $b_x \sim \text{Bernoulli}(p_{noise})$. The perturbed query batch $\tilde{B}^{i}$ is constructed as:
\begin{equation}
\tilde{x} = x + b_x \cdot \epsilon, \quad \forall x \in B^{i}
\end{equation}
where $\epsilon \sim \mathcal{N}(0, \sigma^2I)$ represents distinct Gaussian noise sampled for each query. Note that here $x$ refers to the embedding vector of the query (encoded by the embedding model). This acts as implicit data augmentation to smooth the decision boundary, preventing overfitting in data-scarce scenarios and improves generalization to unseen queries. The inference network $\mathcal{E}_{\phi}$ encodes the contexts into a latent task variable $z$ (line 6), which conditions the policy $\pi_{\theta}$. The policy then processes the batch $\tilde{B}^{i}$ to compute the routing objective (line 7). Finally, the accumulated gradients are used to update the parameters of both networks (lines 9-10). The policy is updated to maximize routing performance, while the inference network is simultaneously updated by backpropagating the policy objective through the entire computational graph. This step trains $\mathcal{E}_\phi$ to produce latent variables $z$ that are maximally informative for the policy, thereby directly linking the quality of the inference to the final routing performance.

\section{Experiments}
\subsection{Experiment Setup}

\begin{table*}[t]
  \centering 
  \setlength{\tabcolsep}{1mm} 
  \small
  \begin{tabular*}{\linewidth}{@{\extracolsep{\fill}}l ccc ccc ccc} 
    \toprule
    \multirow{2}{*}{\textbf{Method}} & \multicolumn{3}{c}{\textbf{RouteLLM dataset}} & \multicolumn{3}{c}{\textbf{MATH}} & \multicolumn{3}{c}{\textbf{Magicoder dataset}}\\
    \cmidrule(lr){2-4} \cmidrule(lr){5-7} \cmidrule(lr){8-10}  
    & HV $\uparrow$  & IGD $\downarrow$ & AUC $\uparrow$ & HV $\uparrow$  & IGD $\downarrow$  & AUC $\uparrow$ & HV $\uparrow$  & IGD $\downarrow$ &AUC $\uparrow$ \\
    \midrule
    Oracle & 0.9242  & 0  & 2.701 & 0.8697 & 0 & 0.5920 & 0.8514 & 0 & 0.4021 \\
    \midrule
    GraphRouter & \underline{$0.7755$}  & \underline{$0.1348$} & \underline{$2.493$} & $0.6134$ & $ 0.1691$ & $\underline{0.5416}$ & $0.7006$ &$0.1281$ & \underline{$0.3885$}\\
    SW ranking & $0.4778$ & $0.3446$& $1.560$  & $0.5820$ & $ 0.2061$& $0.4186$ & $0.5883$ & $0.2031$ & $0.3127$\\
    Matrix factorization & $0.6590$ & $0.2000$ & $2.203$ & $0.6060$ & $0.1885$& $0.4974$ & $0.6007$ & $0.1901$ & $0.3509$\\
    Avengers-Pro & $0.6942$  & $0.1716$& $2.276$ & \underline{$0.6269$} & \underline{$0.1615$} & $0.5392$& \underline{$0.7175$} & \underline{$0.1081$} & $0.3860$\\
    MetaRouter (ours) & $\mathbf{0.8437}$ & $\mathbf{0.0801}$ & $\mathbf{2.583}$ & $\mathbf{0.6716}$ & $\mathbf{0.1419}$& $\mathbf{0.5454}$ & $\mathbf{0.7536}$ & $\mathbf{0.0829}$ & $\mathbf{0.3903}$\\
    \bottomrule
  \end{tabular*}
  \caption{Performance on in-distribution (ID) tasks. We highlight the best results in bold, and the second-best results with an underline. Note that AUC values are scaled by 100 ($\times 10^{-2}$) for better readability.} 
  \label{tab:results} 
\end{table*}

\begin{table*}[t]
  \centering 
  \setlength{\tabcolsep}{1mm}
  \small
  \begin{tabular*}{\linewidth}{@{\extracolsep{\fill}}l ccc ccc ccc}
    \toprule
    \multirow{2}{*}{\textbf{Method}} & \multicolumn{3}{c}{\textbf{AlpacaEval}} & \multicolumn{3}{c}{\textbf{Omni-MATH}} & \multicolumn{3}{c}{\textbf{FullStackBench}}\\
    \cmidrule(lr){2-4} \cmidrule(lr){5-7} \cmidrule(lr){8-10}
    & HV $\uparrow$  & IGD $\downarrow$ & AUC $\uparrow$ & HV $\uparrow$  & IGD $\downarrow$  & AUC $\uparrow$ & HV $\uparrow$  & IGD $\downarrow$ &AUC $\uparrow$ \\
    \midrule
    Oracle & $0.8411$ & 0 & $6.120$ & $0.7548$ & 0 & $1.193$ & $0.8667$ & 0 & $0.2345$\\
    \midrule
    GraphRouter & \underline{$0.6697$} & $\underline{0.1212}$ & $\underline{5.520}$ & $0.5093$ & $0.1827$ & $1.089$ & $0.6156$ & $0.1656$ & $0.2180$\\
    SW ranking & $0.6049$ & $0.2198$& $4.252$ & $0.5254$ & $0.2406$& $0.794$ & \underline{$0.6661$} & $0.1434$ & $0.2166$\\
    Matrix factorization & $0.6243$ & $0.1600$ & $4.978$ & $\mathbf{0.6260}$ & $\mathbf{0.0949}$& $1.080$ & $0.6385$ & $0.1543$ & $0.2152$\\
    Avengers-Pro & $0.6259$  & $0.1497$& $5.426$ & $0.4383$ & $0.2325$& \underline{$1.101$} & $0.6548$ & \underline{$0.1379$}& \underline{$0.2185$} \\
    MetaRouter (ours) & $\mathbf{0.6778}$ & $\mathbf{0.1176}$ & $\mathbf{5.949}$ & $\underline{0.5667}$ & \underline{$0.1353$}& $\mathbf{1.115}$ & $\mathbf{0.6678}$ & $\mathbf{0.1334}$& $\mathbf{0.2202}$\\
    \bottomrule
  \end{tabular*}
  \caption{Performance on out-of-distribution (OOD) tasks. We highlight the best results in bold, and the second-best results with an underline. Note that AUC values are scaled by 100 ($\times 10^{-2}$) for better readability.} 
  \label{tab:generalization} 
\end{table*}

\paragraph{Datasets.} We evaluate our method on three tasks: hybrid question answering (QA), code generation, and mathematical reasoning. For each task, we use one dataset to assess in-distribution (ID) performance and another to test out-of-distribution (OOD) performance: RouteLLM \cite{routellm} and AlpacaEval \cite{alpaca_eval} for Hybrid QA; Magicoder \cite{magicoder} and FullStackBench \cite{liu2024fullstackbenchevaluatingllms} for code generation; and MATH \cite{MATH} and Omni-MATH \cite{omnimath} for mathematical reasoning. For efficiency and due to the large size of the original dataset, we primarily selected difficult queries where small models are less likely to perform as well as large models. The relevant statistics are summarized in Supplementary Material.

\paragraph{Candidate LLM.} For Hybrid QA, as the original dataset already provides responses from gpt-4-1106-preview \cite{gpt4technicalreport} and Mixtral 8x7B \cite{mixtralexperts}, we directly use them respectively as the large model and the small model. For other datasets, we use DeepSeek-v3 \cite{deepseekai2025deepseekv3technicalreport} as the large model and Qwen2.5-1.5B \cite{qwen2technicalreport} as the small model. Given that LLM judges have shown a high correlation with human judgment \cite{alpacafarm}, we use the widely-adopted LLM judge to measure the response quality \cite{routellm}. We also test another quality metric BART score \cite{bartscore} in Section~\ref{sec:sensitivity}. Pricing information for the open-source models is referenced from  TogetherAI \cite{TogetherAI_Pricing_2025}.

\paragraph{Baselines.} We compare MetaRouter with the following strong baselines: \emph{GraphRouter} \cite{graphrouter}, which routes queries via a graph framework requiring separate training for different preferences; \emph{Avengers-Pro} \cite{avengerspro,avengers}, which routes based on embedding-based clustering and a cluster-wise weighted score; and two methods from \cite{routellm}, the training-free \emph{Similarity-weighted (SW) ranking} and the recommended method \emph{Matrix Factorization}, both using thresholds to control trade-offs. We also include \emph{Oracle} to demonstrate the performance upper bound.

\paragraph{Implementation Details.} Following common designs \cite{graphrouter,avengerspro}, we instantiate the score function as $s(o)=w\cdot q(o)-(1-w)\cdot p(o)$, where $q(o)$ and $p(o)$ denote the normalized quality and price, respectively. We also test more complex score function in Section~\ref{sec:sensitivity}. For all routers, we set ten different levels of preference for balancing performance and costs. We adopt all-mpnet-base-v2 \cite{mpnet} as the embedding model to encode queries. Network architectures and other details are provided in the Supplementary Material.

\paragraph{Metrics.} We use two standard metrics in multi-objective optimization: Hypervolume (HV) \cite{HV} for overall quality, and Inverted Generational Distance (IGD) \cite{IGD} for convergence, using the Oracle's solution as the true Pareto front. To complement these geometric metrics, we report the Area Under the Cost-Performance Curve (AUC), which quantifies the router's global efficiency and robustness across the cost spectrum.

\subsection{Main Results}
\begin{figure}[t]
\centering
\includegraphics[width=0.9\columnwidth]{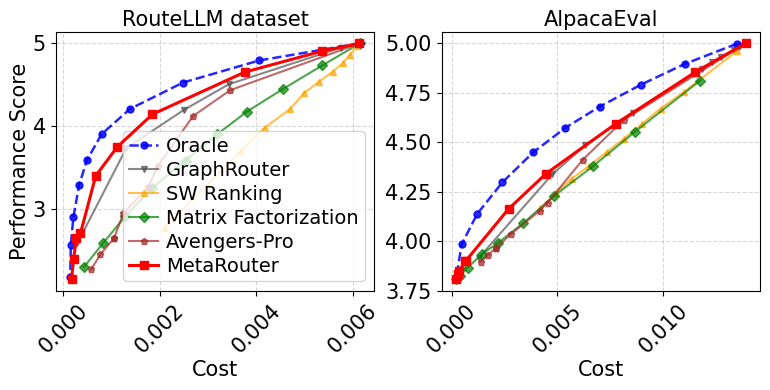} 
\caption{Performance on Hybrid QA task.  The left and right figures show performance on the ID and OOD datasets, respectively. MetaRouter demonstrates better performance than other methods.}
\label{fig:hybridQA}
\end{figure}

Table \ref{tab:results} displays the in-distribution (ID) performance of all methods. As shown, MetaRouter consistently outperforms the baseline methods across all metrics on all ID datasets. For example, it improves the HV score by approximately 8.8\% and reduces the IGD score by 40.6\% compared to the second-best method on the RouteLLM dataset. Furthermore, MetaRouter achieves the highest AUC scores across all tasks. This suggests that MetaRouter is highly effective when the test data distribution matches the training distribution.

Table \ref{tab:generalization} evaluates out-of-distribution (OOD) generalization. MetaRouter demonstrates strong generalization capabilities, achieving the best performance across all metrics on AlpacaEval and FullStackBench. On Omni-MATH, while Matrix factorization achieves slightly better HV and IGD scores, MetaRouter outperforms all baselines in terms of AUC. Overall, MetaRouter maintains robust performance against baselines. We also observed a noticeable gap between the oracle and all methods in Omni-MATH and FullStackBench, which can be attributed to the limited size of the training dataset.

Figure \ref{fig:hybridQA} provides a visual example of the router's performance on both ID and OOD datasets for the Hybrid QA task. Other figures are available in the Supplementary Material.

\subsection{Context Analysis}
We need to collect contexts to infer preferences in the adaptation process. Obviously, the quantity of context significantly impacts performance. If many contexts are required for accurate inference, this method will impose a significant burden on the user, which is unacceptable. In this section, we want to answer the following question: \emph{How many preference contexts $c$ does the context encoder need to infer a reliable $z$?}

We conducted experiments on the RouteLLM dataset. To intuitively measure the impact of context quantity, we defined average task accuracy (ATA). This metric is the average accuracy computed over all test samples from all tasks. A prediction is deemed correct if our router's output matches the output from the oracle. Formally, ATA is defined as:
\begin{equation}
\text{ATA} = \frac{1}{T \cdot D_{\text{test}}} \sum_{j=1}^{T} \sum_{i=1}^{D_{\text{test}}} \mathbb{I}(a_{\text{MetaRouter}}^{ji} = a_{\text{Oracle}}^{ji}),
\end{equation}
where $D_\text{test}$ is the number of samples in the test set and $T$ is the total number of tasks.

\begin{figure}[t]
\centering
\includegraphics[width=0.77\columnwidth]{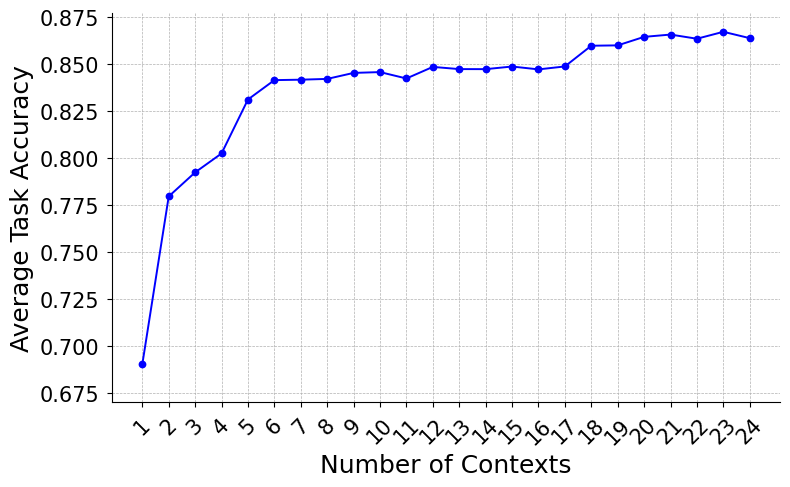} 
\caption{Effect of the number of contexts on MetaRouter's performance on the RouteLLM dataset. High performance can be achieved with only a small number of contexts.}
\label{fig:contexts}
\end{figure}

As shown in Figure \ref{fig:contexts}, the experimental results indicate that only a few (around 6) contexts are needed to achieve strong performance. After that, accuracy continues to rise at a slower pace and appears to plateau around 0.86 after 18 contexts. MetaRouter reaches a saturation point where it has sufficient information for optimal performance.

\subsection{Generalization and Scalability}
\label{sec:gen_scale}

\begin{figure}[t]
    \centering
    \includegraphics[width=0.9\linewidth]{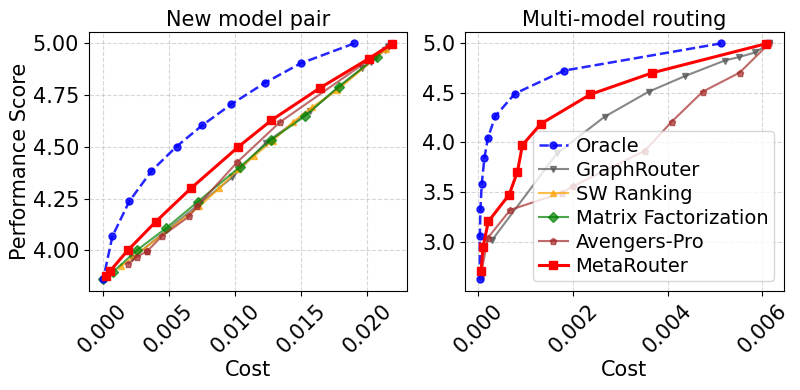} 
    \caption{\textbf{Generalization and Scalability Analysis.} \textbf{(Left)} Performance of MetaRouter when transferring to a new model pair (Claude 3 Opus and Llama 3 8B) without retraining. \textbf{(Right)} Performance when scaling to a multi-model routing scenario with five LLMs.}
    \label{fig:gen_scale}
\end{figure}

\paragraph{Generalization to New Model Pair.} To evaluate generalization, we tested the router on AlpacaEval using a new model pair (Claude 3 Opus \cite{Claude3Opus} and Llama 3 8B \cite{llama3modelcard}) instead of the original training pair (GPT-4 \cite{gpt4technicalreport} and Mixtral-8x7B \cite{mixtralexperts}), without retraining. As shown in Figure \ref{fig:gen_scale} (left), MetaRouter consistently outperforms baselines, demonstrating strong generalization capabilities.

\paragraph{Scalability to Multi-Model Routing.} We further extended our framework to a multi-model setting involving five candidates on the RouteLLM dataset: Mixtral-8x7B, Llama 3 8B, Yi 34B \cite{ai2024yi}, Llama 3 70B, and GPT-4. To adapt to this scenario, we maintain the pairwise comparison mechanism during the preference adaptation phase by sampling response pairs from the candidate pool. Crucially, our reward formulation (Eq.~\eqref{eq:reward}) naturally scales to this setting, requiring no architectural changes other than adjusting the policy's output dimension to 5. Note that baselines strictly limited to binary settings (SW Ranking and Matrix Factorization) were excluded. Figure \ref{fig:gen_scale} (right) shows that MetaRouter significantly outperforms other baselines, indicating that our formulation scales naturally to larger action spaces.

\subsection{Sensitivity Analysis}
\label{sec:sensitivity}

\begin{figure}[t] 
    \centering
    \includegraphics[width=0.9\linewidth]{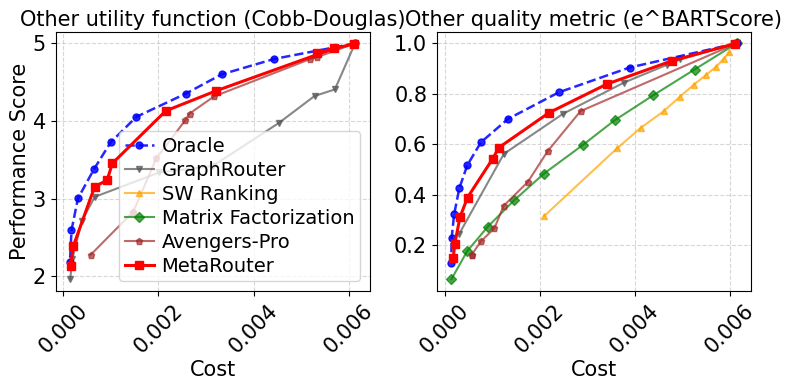}
    \caption{\textbf{Sensitivity Analysis.} \textbf{(Left)} Performance comparison under the non-linear Cobb-Douglas utility function. MetaRouter (red) remains closest to the Oracle (blue). \textbf{(Right)} Evaluation using transformed BART scores as the quality metric. The proposed method consistently outperforms baselines across varying cost constraints.}
    \label{fig:sensitivity}
\end{figure}

\paragraph{Utility Functions.} To assess adaptability to complex user preferences, we employed the non-linear Cobb-Douglas utility function~\cite{cobb1928theory}, defined as $s(o) = q(o)^{1-w} \cdot p(o)^{-w}$. We exclude threshold-based baselines as they do not explicitly optimize utility. Figure~\ref{fig:sensitivity} (Left) shows that MetaRouter maintains its performance advantage, confirming its robustness to diverse preference structures.
\paragraph{Quality Metric for Responses.} We substituted the LLM judge with the BART score~\cite{bartscore}. To align this metric with our utility formulation, we transformed the original log-probability scores into the $[0, 1]$ range using an exponential function($e^{\text{BART Score}}$). Figure \ref{fig:sensitivity} (Right) confirms that MetaRouter consistently outperforms the baselines, regardless of the metric employed.

\subsection{Ablation Study}

We conducted ablation studies to validate the contribution of Entropy Regularization (ER), the tanh operation in reward design, the Gated Residual Mechanism (GRM), and Noise Injection (NI). The proposed MetaRouter yields the best performance with an HV of $0.8437$. The removal of ER and NI decreases the HV to $0.8236$ and $0.8233$, respectively. Furthermore, the IGD metric worsens from $0.0801$ to $0.0935$ when GRM is excluded (replacing it with a direct concatenation of $z$ and $x$). This consistent performance drop across different metrics confirms the effectiveness of our design. Please refer to the Supplementary Material for the complete table.

\section{Conclusion}
In this paper, we study the problem of balancing cost and performance in LLM applications. To this end, we introduce perceptive LLM routing, a novel user-centric paradigm. To realize this paradigm, we present MetaRouter which is the first framework to leverage meta-learning for LLM routing. Experimental results demonstrate that MetaRouter significantly outperforms strong baselines on both in-distribution and out-of-distribution tasks. Furthermore, it exhibits strong generalization to unseen model pairs and scalability to multi-model routing. Future work will aim to expand the scope of user preferences to include other critical dimensions such as latency and privacy.


\section*{Acknowledgements}
This work is supported by National Natural Science Foundation of China (Project 62502412), Guangdong Province (Project 2024QN11X097), and HKUST-HKUST(GZ) Cross-campus Collaborative Research Scheme under the “1+1+1” Joint Funding Program (G081).

\bibliographystyle{named}
\bibliography{ijcai26}

\end{document}